\documentclass[sigconf]{acmart}
\usepackage{epsfig}
\usepackage{graphicx}
\usepackage{amsmath}
\usepackage{multirow}
\usepackage{subfigure}
\usepackage{balance}

\newcommand{\tabincell}[2]{\begin{tabular}{@{}#1@{}}#2\end{tabular}} 
\usepackage{amsfonts}
\usepackage{mathrsfs}
\usepackage{subfigure}
\usepackage{wrapfig}
\usepackage{makecell}

\usepackage[switch]{lineno}  %

\usepackage{paralist} % squeezed itemized lists

\newenvironment{packed_itemize}{
	\begin{itemize}
		\setlength{\itemsep}{0pt}
		\setlength{\parskip}{0pt}
		\setlength{\parsep}{0pt}
	}{\end{itemize}}

\AtBeginDocument{%
	\providecommand\BibTeX{{%
			\normalfont B\kern-0.5em{\scshape i\kern-0.25em b}\kern-0.8em\TeX}}}

\copyrightyear{2021}
\acmYear{2021}
\setcopyright{acmcopyright}\acmConference[MM '21]{Proceedings of the 29th ACM International Conference on Multimedia}{October 20--24, 2021}{Virtual Event, China}
\acmBooktitle{Proceedings of the 29th ACM International Conference on Multimedia (MM '21), October 20--24, 2021, Virtual Event, China}
\acmPrice{15.00}
\acmDOI{10.1145/3474085.3475243}
\acmISBN{978-1-4503-8651-7/21/10}

\begin{document}
	\fancyhead{}
	\title{Revisiting Mid-Level Patterns \\ for Cross-Domain Few-Shot Recognition}
	
	\author{Yixiong Zou$^{1,3}$, Shanghang Zhang$^2$, Jianpeng Yu$^3$, Yonghong Tian$^{1*}$, Jos\'e M. F. Moura$^3$}
	\thanks{$^*$ indicates corresponding author.}
	\affiliation{%
		\institution{Peking University$^1$, University of California, Berkeley$^2$, Carnegie Mellon University$^3$}
		\country{}
	}
	\email{{zoilsen, yhtian}@pku.edu.cn, shz@eecs.berkeley.edu,  {jianpeny, moura}@andrew.cmu.edu}

	\renewcommand{\shortauthors}{Yixiong Zou, et al.}
	
	%%%%%%%%% ABSTRACT
	\begin{abstract}
	Existing few-shot learning (FSL) methods usually assume base classes and novel classes are from the same domain (in-domain setting). However in practice, it may be infeasible to collect sufficient training samples for some special domains to construct base classes. To solve this problem, cross-domain FSL (CDFSL) is proposed very recently to transfer knowledge from general-domain base classes to special-domain novel classes. Existing CDFSL works mostly focus on transferring between near domains, while rarely consider transferring between distant domains, which is even more challenging.
	In this paper, we study a challenging subset of CDFSL where the novel classes are in distant domains from base classes, by revisiting the mid-level features, which are more transferable yet under-explored in main stream FSL work. To boost the discriminability of mid-level features, we propose a residual-prediction task to encourage mid-level features to learn discriminative information of each sample. Notably, such mechanism also benefits the in-domain FSL and CDFSL in near domains. Therefore, we provide two types of features for both cross- and in-domain FSL respectively, under the same training framework. Experiments under both settings on six public datasets, including two challenging medical datasets, validate the rationale of the proposed method and demonstrate state-of-the-art performance. Code will be released\footnote{https://pkuml.org/resources/code.html}.
\end{abstract}





\maketitle

\section{Introduction}
\label{sec: introduction}

\begin{figure}[t]
	\centering
	\subfigure{\includegraphics[width=0.9\columnwidth, height=0.55\columnwidth]{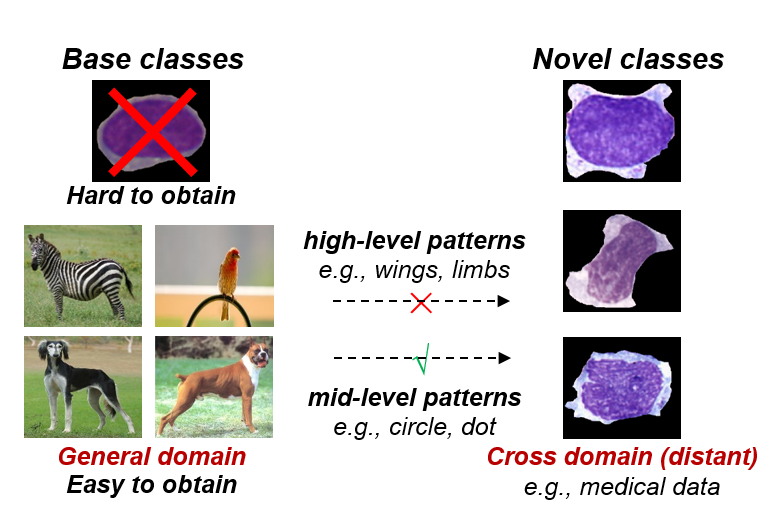}} \\
	\vspace{-0.35cm}
	\subfigure{\includegraphics[width=1.0\columnwidth]{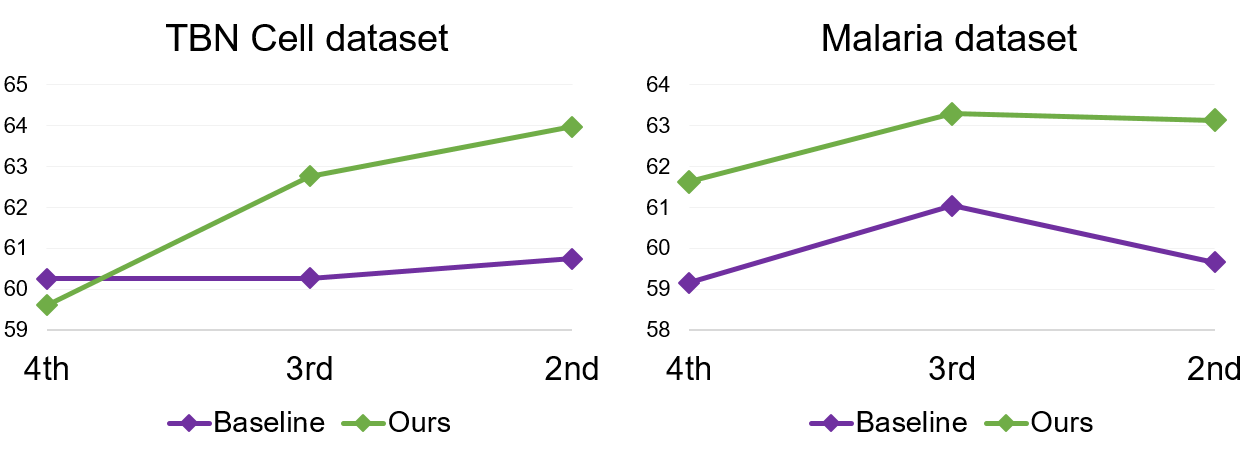}}
	\vspace{-0.55cm}
	%\caption{Few-shot learning is useful for domains where data is hard to obtain (e.g., medical data), which means it may also be infeasible to construct a base-class set with sufficient data from that domain. To transfer knowledge from easy-to-obtain base classes in the general domain to novel classes in the distant-domain, we revisit mid-level patterns that are more transferable than high-level patterns.}
	\caption{Top: Samples in the general domain are easy to obtain, while they may be hard to obtain in special domains (e.g. medical data) which could be distant from the general domain. To transfer knowledge from easy-to-obtain base classes in the general domain to novel classes in the distant domain (a challenging subset of cross-domain few-shot recognition), we revisit mid-level patterns that are more transferable than high-level patterns. Bottom: Quantitative evaluation of features from different blocks of ResNet when transferring base-class trained model to distant cross-domain datasets, where mid-level features (3rd and 2nd blocks) could show better performance compared with the high-level features (4th block, i.e., the last layer).}
	\vspace{-0.3cm}
	\label{fig: motivation}
\end{figure}

Few-shot learning (FSL)~\cite{Vinyals2016Matching} has been proposed recently to recognize objects in novel classes given only few training samples, with knowledge transferred from base classes (classes with sufficient training samples).
%Current FSL works~\cite{Vinyals2016Matching, qiao2017few} mainly focus on the recognition/classification task. 
%Existing FSL work~\cite{Vinyals2016Matching, qiao2017few} usually assumes there is a large dataset of base classes which contain sufficient training data \textit{from the same domain} of the novel classes, and there are no classes in overlap with the novel classes (in-domain setting).
Existing FSL works~\cite{Vinyals2016Matching, qiao2017few} usually assume the in-domain setting, where base classes and novel classes are from the same domain.
%, imitating humans' behavior of transferring prior knowledge from base classes to novel classes.
However, such setting may not stand in practice, because for domains where data is hard to obtain, it may be infeasible to collect sufficient training samples from them to construct the base classes either, as shown in Fig.~\ref{fig: motivation}. To solve this problem, very recently, cross-domain FSL (CDFSL)~\cite{DBLP:journals/corr/abs-1904-04232,tseng2020cross} has been proposed to handle a more realistic setting where data from the general domain (which is easier to collect~\cite{DBLP:journals/corr/abs-1904-04232}, e.g., ImageNet~\cite{deng2009imagenet}) are sampled as base classes, while data from other domains are defined as novel classes. 
Compared with the general domain, the novel-class domains may contain semantic shift (general-domain to birds~\cite{DBLP:journals/corr/abs-1904-04232}), style-shift (natural images to pencil-paintings~\cite{zhao2020domain}), or both~\cite{guobroader} (general-domain to medical microscopic images, as shown in Fig.~\ref{fig: motivation}).
The novel-class domains may vary from being close to being distant against the base-class domain~\cite{guobroader}, because no assumptions could be made about what novel classes would appear and no one could enumerate all possible classes in base classes.
%Such CDFSL problem requires knowledge transfer for cross domains from close to distant~\cite{guobroader}, which still remains a great challenge.

However, existing CDFSL works~\cite{DBLP:journals/corr/abs-1904-04232,tseng2020cross} mostly focus on the transferring between domains that are close to each other, while rarely consider that for distant domains.
%, for which their methods always suffer. which is even more challenging.
For instance, some specialized domains such as medical domains usually lack labeled training samples and are very different from the general domain. It is beneficial while challenging to transfer knowledge from general-domain to facilitate recognition of novel classes in these specialized domains.
Therefore in this paper, we aim to solve a more challenging subset of the CDFSL problem where base classes and novel classes are from distant domains, termed as \textit{distant-domain FSL} for abbreviation.
%Moreover, current works merely intuitively choose cross-domain novel classes, which lacks quantitative measure of the domain distance.
%Therefore, we also use the Proxy-A-Distance (PAD)~\cite{ganin2016domain,ben2007analysis} to quantitatively measure domain distances to facilitate the choice of distant domains. 
Moreover, to facilitate the definition of distant domains, we also use the Proxy-A-Distance (PAD)~\cite{ganin2016domain,ben2007analysis} to quantitatively measure domain distances (as shown in Tab.~\ref{tab:datasets} and section~\ref{sec: pad}).

%and provide two medical datasets\footnote{TBN Cell~\cite{pansombut2019convolutional}: http://mcs.sat.psu.ac.th/dataset/dataset.zip\\Malaria Cell~\cite{rajaraman2018pre}: https://lhncbc.nlm.nih.gov/publication/pub9932}, which may be practically useful for medical multimedia applications due to the scarcity of medical data. 

To address distant-domain FSL, the model should learn transferable patterns from general-domain base classes and transfer them to distant-domain novel classes. 
Much work~\cite{yosinski2014transferable} on transfer learning suggests features from shallower (mid-level) layers are more transferable than those from deeper layers.
Intuitively, as shown in Fig.~\ref{fig: motivation} (top), high-level patterns from the general domain, such as wings and limbs, can hardly be transferred to distant-domain novel classes, while mid-level patterns, such as circle and dot, are easier to be transferred.
Quantitatively, as shown in Fig.~\ref{fig: motivation} (bottom), mid-level features from the third and second blocks of ResNet~\cite{he2016deep} could show better performance than the high-level features from the forth block (i.e., the last block) when transferring base-class trained model to distant-domain datasets\footnote{To avoid ambiguity, we denote the feature/pattern of the last layer of the backbone network as high-level feature/pattern, and denote the feature/pattern of the layers other than the first and the last layer as the mid-level feature/pattern}.
The utilization of mid-level features has been widely explored in the research on transfer learning~\cite{long2015learning}, yet it is far from being well explored in FSL.
Therefore, in this paper, we revisit mid-level features to learn transferable and discriminative mid-level features for distant-domain FSL.
%In this paper, we try to utilize mid-level (shallower) features to boost the weakly-related novel-class recognition.

Although mid-level features are more transferable than high-level features, they may not be discriminative enough.
To boost their discriminability, during the base-class training, we design a residual-prediction task to encourage mid-level features to learn the discriminative information of each sample.
The insight is that we assume each class has its unique character that could not be easily described by high-level patterns from other classes, while mid-level patterns can be more effective to describe it. Such unique character provides information to learn more discriminative mid-level features.
Intuitively, for example in Fig.~\ref{fig: motivation}, to describe the unique character of zebra, zebra stripes, with knowledge from dogs, it is hard to use high-level patterns (e.g., high-level semantic part~\cite{tokmakov2019learning}) from dogs, while it is much easier to use mid-level patterns such as stripe to describe it, which indicates such unique character provides suitable information to facilitate the learning of more discriminative mid-level features.
Moreover, as an example, such stripe-like pattern could help the medical analysis~\cite{raghu2019transfusion}, i.e., distant-domain recognition.
%which indicates such unique character contains information suitable for mid-level features to learn.
%to describe zebra with knowledge learned from dogs, it is easy to transfer high-level patterns such as feet, tail to zebra. 
%To learn such information, 
Specifically, we first extract features for base-class samples with the backbone network being trained by classifying such sample into $N$ base classes.
Then, for each training sample, we use high-level patterns from other $N-1$ base classes to reconstruct the extracted feature, and obtain the residual feature by \textbf{removing} the reconstructed feature from the extracted feature.
%, outputting a discriminative residual feature, 
Such residual feature contains the discriminative information of this sample that is suitable for the mid-level features to learn.
Finally, we choose all layers in the backbone network other than the first layer and the last layer to be the mid-layer set, layer-wisely learn weights for all the corresponding mid-level features, and force the weighted combined mid-level features to predict the discriminative residual features, which encourages mid-level features to be discriminative.

Note that although we aim at boosting the distant-domain FSL, our method is also effective for in-domain FSL and CDFSL in near domains. 
The base-class training process designed above is a pseudo-novel-class training strategy, which views the current training sample as a pseudo-novel-class sample, providing simulated in-domain novel-class data, and views the other classes as pseudo-base classes.
As adequate information is provided when classifying such sample into base classes, its feature can be viewed as the ground truth for the pseudo-novel training process,
and we are trying to predict such pseudo-novel features by reconstructing them via high-level patterns from pseudo-base classes.
The lower bound of the reconstruction loss validates our assumption that each class has its unique character (Tab.~\ref{tab:recon_methods}).
On the other hand, by predicting the discriminative pseudo-novel residual features, we are also encouraging the model to have the capability to predict real residual features for the real novel-class samples.
%and encouraging the extracted feature to be decomposed by the high-level and mid-level feature,
Combining the predicted high-level feature and residual feature would output the whole predicted feature for the in-domain and near-domain novel class sample, thus boosting the in-domain FSL and CDFSL in near domains.
%For the distant-domain FSL, we learn discriminative mid-level features via residual-prediction task and use the weighted concatenation of mid-level features as the feature for the novel-class samples.
Therefore, we provide two types of features for both distant-domain and in-domain novel-class recognition respectively, under the same training framework,
according to the quantitative measure of domain distances by the Proxy-A-Distance (PAD)~\cite{ganin2016domain}.
For distant-domain novel classes, we use the weighted concatenation of mid-level features in the candidate mid-layer set as the final feature.
For in-domain or near-domain novel classes, we use all base-class prototypes to perform the high-level feature reconstruction, and combine it with the predicted residual term to be the final feature. 
%Moreover, the backbone-extracted feature is also provided as a simplified feature for the in-domain FSL, which is latently encoded with both the high-level and mid-level features (section~\ref{sec: component}).
Finally, the nearest neighbor classification will be performed for the novel-class recognition for all settings.

In all, our contributions can be summarized as follows:
\vspace{-0.1cm}
\begin{packed_itemize}
	
	%\item We focus on a more challenging problem: distant-domain FSL, which is especially important for applications such as medical analysis due to the scarcity of data, and utilize PAD to quantitatively measure the domain distances. %We go a step further to provide a more challenging scenario, distant-domain FSL, with two medical classification datasets, and use PAD to quantitatively measure the domain distances.
	
	\item To solve CDFSL in distant domains, we revisit mid-level features to explore their transferability and discriminability, which is seldom studied in the main stream FSL work.
	
	%\item To boost the discriminability of mid-level features, we propose a residual-prediction task to simulate the usage of mid-level features to describe novel classes, which computes the residual between the extracted pseudo-novel-class feature and the high-level reconstructed feature. The residual term is predicted by a dynamic mid-layer-selection module with linear transforming layers on mid-level features.
	\item To enhance the discriminability of mid-level features, we propose a residual-prediction task to explore the unique character of each class.
	%which also benefits the in-domain FSL
	
	\item Our method is effective for not only the distant-domain FSL but also the in-domain FSL and near-domain CDFSL with different types of descriptive features. Experiments under both settings on six public datasets, including two challenging medical datasets, demonstrate state-of-the-art performance. Code will be released.
	%validate the effectiveness of our method and
	
\end{packed_itemize}
%%\vspace{-0.15cm}

\section{Related Work}

\textbf{Few-shot learning} methods can be roughly grouped into embedding based method~\cite{Vinyals2016Matching,yang2018learning,garcia2017few,zou2020compositional,zou2018hierarchical,zou2020annotation},
%which learns an embedding space where samples from the same classes are close while those from different classes are distant,
meta-learning based method~\cite{andrychowicz2016learning,munkhdalai2017meta,finn2017model,zou2020adaptation},
%which learns a meta-learner to guide the adaptation on novel classes, 
and hallucination based method~\cite{hariharan2017low,wang2018low}.
%which hallucinates samples to augment the training data.
%The pseudo-novel-class strategy is also adopted in ~\cite{gidaris2018dynamic}, which inspires us to take such strategy for novel-class simulation.
The pseudo-novel-class strategy is also adopted in ~\cite{gidaris2018dynamic}.
Very recently, some works~\cite{tseng2020cross, DBLP:journals/corr/abs-1904-04232, triantafillou2019meta, chen2020new, zhao2020domain} studied the problem of cross-domain FSL, which train the model on general-domain classes and evaluate it on novel classes from other domains.
%such as CUB~\cite{wah2011caltech} and pencil-paintings~\cite{zhao2020domain}. % (i.e.,  \textit{mini}ImageNet~\cite{snell2017prototypical}) 
%Most works~\cite{DBLP:journals/corr/abs-1904-04232, triantafillou2019meta, chen2020new} focus on validating baseline methods on this task or proposing datasets. 
\cite{tseng2020cross} proposed to insert affine transformations sampled from the Gaussian distribution to intermediate layers to help the generalization.  \cite{zhao2020domain} proposed to utilize the domain adversarial adaptation mechanism to handle the style shift problem. 
%The above settings contain novel classes with either semantic shift (general-domain classes to fine-grained bird classes) or style shift (nature images to pencil-paintings), which we summarize as \textit{weakly-related} to base classes. 
We also study the problem of cross-domain FSL, and focus on a more challenging subset, i.e., distant-domain FSL.

%\vspace{0.1cm}
\noindent\textbf{Transferability} of deep networks has been researched in the field of transfer learning~\cite{yosinski2014transferable}, which shows an decreasing trend of transferability when going deeper into the deep network. Such phenomenon has also been applied in applications such as ~\cite{long2015learning,zhong2016leveraging,zhang2018person}.
%that adapts the network to the target domain with multiple layers within the backbone network. 
Some works~\cite{li2019large,liu2019learning,liu2019prototype} in FSL utilize features of multiple appended layers to handle the hierarchy of classes. 
The only work makes use of mid-level features, to the best of our knowledge, is ~\cite{huang2019all}, which directly applied mid-level features to the classification.%, and is implemented by us for comparison.
%However, the method of ~\cite{huang2019all} is carried on the fine-grained classification from the same domain (i.e., train on CUB and test on CUB), which is not designed for the problem we are facing.
%As far as we are concerned, we are the first to apply mid-level features within the backbone network in handling the problem of transferring knowledge to cross-domain novel classes in the field of few-shot recognition.
In all, the usage of mid-level features is far from being well explored in FSL yet, which we revisit in this paper to boost both the distant-domain and the in/near-domain FSL.

%Similar form of prototype-based-estimation is applied in ~\cite{gidaris2018dynamic,rusu2019meta,shu2018odn}. In ~\cite{gidaris2018dynamic} and ~\cite{rusu2019meta}, prototypes are quired with a complicated key-value mechanism, and meta-learning is utilized to learn an initialization based on the averaged prototypes. In ~\cite{shu2018odn}, weights in the fully connected layers are queried to form the weights for the newly coming classes. Compared with them, our method focuses not only on learning a good estimation of novel-class features but also on learning an improved extracted feature implicitly encoded with neighboring prototypes, by means of the estimation loss. Moreover, we also segment prototypes and the feature into segments to generate a finer estimation of the feature.

%%\vspace{-0.15cm}
\section{Methodology}

\begin{figure}[t]
	\centering
	\hspace*{-0.3cm}
	\subfigure{\includegraphics[width=1.1\columnwidth]{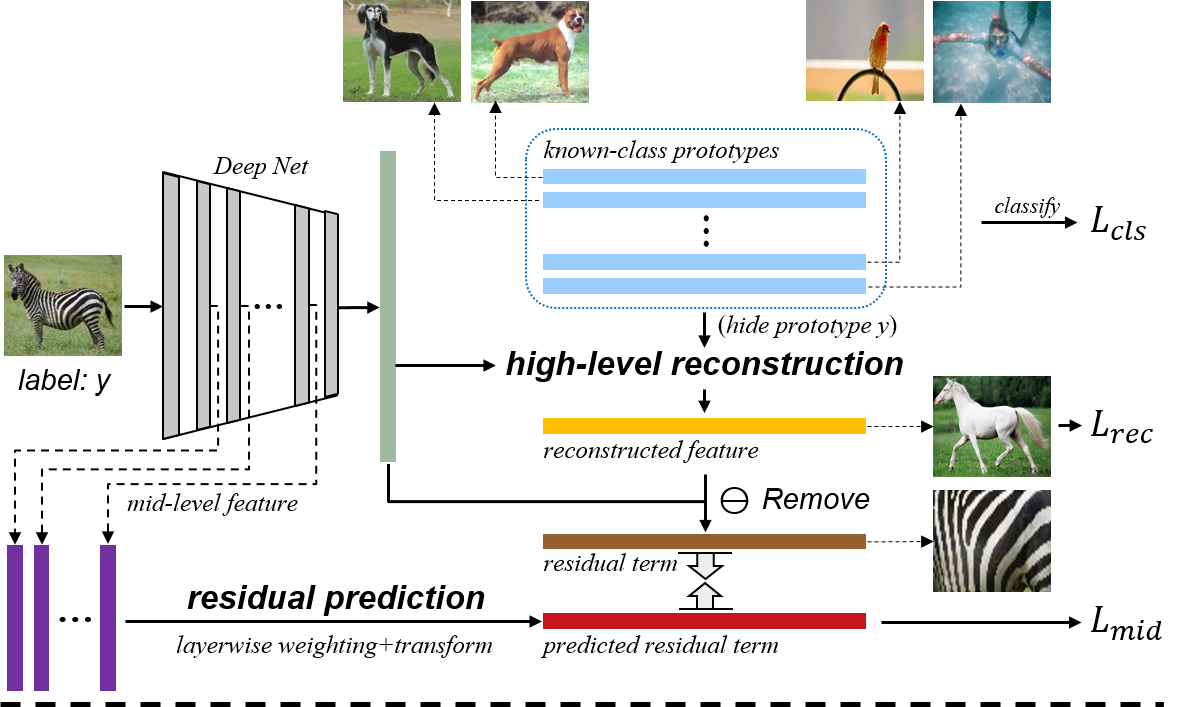}}\vspace{-0.2cm}
	\\
	\subfigure{\includegraphics[width=1.0\columnwidth, height=0.35\columnwidth]{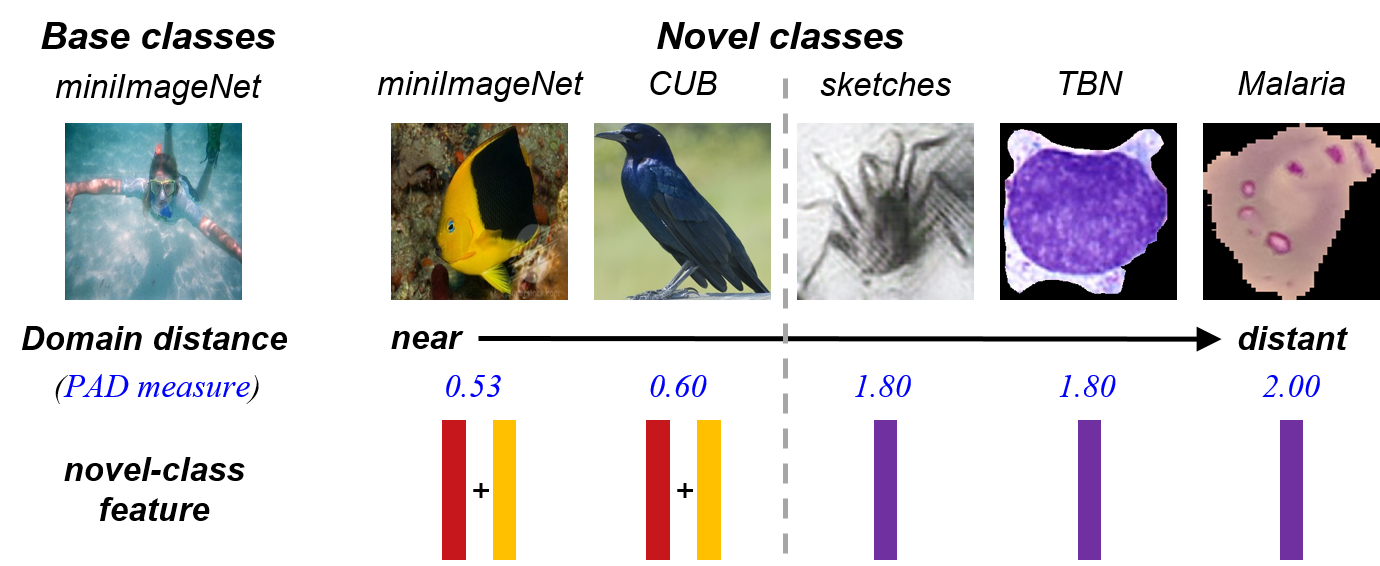}}\vspace{-0.35cm}
	\caption{
		%Framework description. Top: When training on base classes, given a training sample, besides classifying it into N base classes, we first conduct high-level feature reconstruction based on the other N-1 base classes' segmented prototypes. Then the residual term will be calculated as the difference between the extracted feature and the reconstructed feature. Mid-level features from multiple mid-layers will be dynamically weighted to linearly predict the residual term. Such training will benefit both the distant-domain and in-domain FSL, therefore, (Bottom:) while testing on novel classes, we provide two types of features for both in-domain and distant-domain novel-class recognition respectively. 
		Framework. Top: Given a training sample (e.g., zebra) from base classes, besides classifying it into $N$ base classes, we also conduct high-level feature reconstruction based on the other $N-1$ base classes' prototypes (e.g., dogs, birds, human). Then the residual term will be calculated as the difference between the extracted feature and the reconstructed feature (e.g., zebra without stripes, maybe a white horse). Mid-level features from multiple mid-layers will be dynamically weighted to predict the residual term (e.g., stripes).
		%For better understanding, an example of semantic meanings for each vector is plotted in this figure.
		Such training will benefit both the distant-domain and in/near-domain FSL. Bottom: When testing on novel classes, we provide two types of features for both distant-domain and in/near-domain novel classes respectively. 
		%\textit{recon} $\rightarrow$ \textit{reconstruction}, \textit{mid} $\rightarrow$ \textit{mid-features}, \textit{cls} $\rightarrow$ \textit{classify}.
	}\vspace{-0.4cm}
	\label{fig: framework}
\end{figure}

To learn transferable and discriminative mid-level features, we propose a residual-prediction task to explore the unique character of each class, which will benefit both the distant- and in/near-domain FSL.
%Specifically, given an input sample, we first reconstruct the extracted feature by base-class prototypes and calculate the residual term that the high-level patterns fail to represent.
%Then, mid-level features from multiple mid-layers will be dynamically weighted to linearly predict the residual term.
%Such training will benefit both the in-domain and distant-domain FSL.
%Finally we provide two types of features for the in-domain and distant-domain novel-class recognition respectively,
%and the nearest neighbor classification will be performed on novel classes. 
The framework is shown in Fig.~\ref{fig: framework}.

%\begin{figure*}[t]
%	\centering
%	\subfigure{\includegraphics[width=0.85\columnwidth, height=0.46\columnwidth]{framework.png}}\hspace{2.5cm}
%	\subfigure{\includegraphics[width=0.8\columnwidth, height=0.42\columnwidth]{framework_2.png}}%\vspace{-0.3cm}
%	\caption{Framework of our method. Left: While training on base classes, besides the base-class classification, we first conduct high-level feature reconstruction based on the neighboring prototypes of that sample. Then the residual term will be calculated as the difference between the extracted feature and the reconstructed feature. Mid-level features from multiple mid-layers will be dynamically weighted to linearly predict the residual term. Such training will benefit both the in-domain and distant-domain FSL, therefore, (Right:) while testing on novel classes, we provide two types of features for both in-domain and distant-domain novel-class recognition respectively. 
%		%\textit{recon} $\rightarrow$ \textit{reconstruction}, \textit{mid} $\rightarrow$ \textit{mid-features}, \textit{cls} $\rightarrow$ \textit{classify}.
%	}%\vspace{-0.3cm}
%	\label{fig: framework}
%\end{figure*}

\vspace{-0.2cm}
\subsection{Preliminaries}
\label{sec: preliminaries}

Few-shot learning (FSL) aims at recognizing novel classes given only few training samples. Following the setting of current works~\cite{ravi2016optimization}, we are provided with both base classes $\mathcal{C}^{base}$ with sufficient training samples, and novel classes $\mathcal{C}^{novel}$ where only few training samples are available. 
Note that $\mathcal{C}^{base}$ and $\mathcal{C}^{novel}$ are non-overlapping.
The difference between in-domain FSL and cross-domain FSL lies in whether $\mathcal{C}^{base}$ and $\mathcal{C}^{novel}$ are from the same domain~\cite{DBLP:journals/corr/abs-1904-04232}.
Few-shot learning is conducted on the training set (a.k.a. support set) of $\mathcal{C}^{novel}$, and the evaluation is carried on the corresponding testing set (a.k.a. query set). For a fair comparison, current works always conduct a $K$-way $n$-shot evaluation, which means $K$ novel classes $\{C^U_i\}_{i=1}^K$ will be sampled from $\mathcal{C}^{novel}$ with $n$ novel-class training samples $\{x^U_{ij}\}_{j=1}^n$ in each class. For each sampled dataset (i.e., $K \cdot n$ training samples $+$ testing samples, a.k.a. episode), the nearest neighbor classification will be performed, which is represented as 

\vspace{-0.2cm}
\begin{equation}
\small
\hat{y_q} = \mathop{\arg\max}_{y_i} P(y_i|x^U_q)= \mathop{\arg\max}_{i} \frac{e^{s(F(x^U_q), p^U_i)}}{\sum_{k=1}^{K} e^{s(F(x^U_q), p^U_k)} }
\label{eq:MN_P}
\end{equation}
where $F()$ is the feature extractor, $x^U_q$ is the testing sample (a.k.a. query sample), $y_i$ refers to class $C^U_i$, $\hat{y_q}$ is the estimated label for $x^U_q$, $s(,)$ is the similarity function (e.g., cosine similarity), and $p^U_i$ is the estimated prototype for class $C^U_i$, which is typically calculated as $p^U_i = \frac{1}{n} \sum_{j=1}^{n} F(x^U_{ij})$~\cite{snell2017prototypical}. Based on $\hat{y_q}$, the performance will be evaluated on the sampled dataset. Repeat this sampling-evaluation procedure for hundreds of times, the performance of the evaluated model will be obtained.

Before the non-parametric training and testing on novel classes, the model also needs to be (pre-)trained on the base classes~\cite{ravi2016optimization} to learn prior knowledges.
In this work, we utilize the cosine classifier~\cite{li2019large, DBLP:journals/corr/abs-1904-04232} to be our baseline model, which is regarded as a simple but effective baseline. Given $N$ base classes $\{C_i\}_{i=1}^N$, it trains the model by the cross-entropy loss given the input $x$ and its label $y$ as

\vspace{-0.2cm}
\begin{equation}
\small
L_{cls} = -log(P(y|x)) = -log(\frac{e^{\tau {W^c_{y}} f^c(x)}}{\sum_{i=1}^{N} e^{\tau {W^c_{i}} f^c(x)} })
\label{eq:classification loss}
\end{equation}
where $f(x) \in R^{d \times 1}$ is the extracted feature using the backbone $f()$, $W \in R^{N \times d}$ is the parameter for the fully connected (FC) layer,  the superscript $^c$ denotes the vector is $L_2$ normalized ($W^c_{i} = W_{i} / ||W_{i}||_2$, $f^c(x) = f(x)/||f(x)||_2$), and $\tau$ is a pre-defined hyper-parameter. We follow \cite{qiao2017few} to abandon the biases term of the FC layer. As the forward pass of the FC layer is equivalent to the calculation of the cosine similarity of $W^c_{i}$ and $f^c(x)$, this baseline is named the cosine classifier. 
After the training on $\mathcal{C}^{base}$, the backbone will be applied directly as the feature extractor for novel classes (i.e., set $F=f$ in Eq.~\ref{eq:MN_P}), and the nearest neighbor classification will be performed.

\vspace{-0.25cm}
\subsection{Residual-prediction task}
\label{sec: high-level feature estimation}

Although mid-level patterns could be more transferable than high-level ones~\cite{yosinski2014transferable}, they may not be discriminative enough. 
Therefore, to boost the discriminability of mid-level features, we propose a residual-prediction task for the base-class training which encourages mid-level features to learn the discriminative information in each sample.
Intuitively, for example, to describe zebra with knowledge from dogs, it is easy to transfer high-level patterns such as feet, tail to zebra. But for zebra's unique character, zebra stripes, it is hard to transfer high-level patterns (e.g., semantic parts~\cite{tokmakov2019learning}) from dogs, but it is much easier to transfer mid-level patterns such as stripe itself to describe it. 
Also, as an example, such stripe-like pattern could help the medical analysis~\cite{raghu2019transfusion}, i.e., distant-domain recognition.
Inspired by this, we assume every class has its unique character that could not be easily described by high-level patterns from other classes, for which mid-level patterns can be more effective, providing discriminative information suitable for mid-level features to learn.
To improve mid-level features with such information, the residual prediction task can be divided into the following steps as shown in Fig.~\ref{fig: framework} (top): 
we first extract the feature for each base-class sample (e.g., zebra) with the backbone network being trained by the classification loss in Eq.\ref{eq:classification loss}.
Then, for each sample, we design to use high-level patterns from other classes (e.g., dogs, birds, human) to reconstruct the extracted feature (high-level reconstruction), and we \textbf{remove} the reconstructed feature (e.g., zebra without stripe, maybe a white horse) from the extracted feature, outputting a discriminative residual feature (e.g., stripes), which contains the discriminative information for this sample that is suitable for mid-level features to learn.
Finally, we constrain mid-level features to predict such discriminative residual feature, which pushes mid-level features to be discriminative. 
Our method is jointly trained with $L_{cls}$ and the residual-prediction task. Details are in the following.

\vspace{-0.1cm}
\subsubsection{\textbf{High-level Reconstruction}}

Firstly, given a training sample $x$, we use high-level patterns from other base classes to represent (reconstruct) its extracted feature $f(x)$.
Current works~\cite{qiao2017few,gidaris2018dynamic} suggest that the parameters of the base-class FC parameters $W$ could be viewed as the prototypes of base classes, and each row of $W$ (a prototype) contains the overall information of the corresponding class, which refers to the high-level patterns because it exists in the same feature space as that of the backbone's final layer.
Therefore, prototypes are used to reconstruct $f(x)$. 
The prototypes of the other $N-1$ base classes for $x$ is denoted as the prototype set $\{W_i\}_{i \neq y}$ where $y$ is the label of $x$ and ${W}_i \in R^d$ is the same as the corresponding row in the FC parameters $W$.

The reconstruction is based on the feature and prototypes averagely split along the channel axis. For easy understanding, we begin with the situation where no splitting is applied.
Specifically, we use the extracted feature $f(x)$ to apply the nearest neighbor search over $\{W_i\}_{i \neq y}$, and query top $m$ prototypes with the highest cosine similarities to form the neighboring prototype set $\{\widetilde{W}_i\}_{i=1}^m$.
Then, the reconstructed feature is calculated as the mean of all queried prototypes as
$R(x, W) = \frac{1}{m} \sum_{i=1}^{m} \widetilde{W}_i$.

\begin{figure}[t]
	\centering
	\hspace*{-0.3cm}
	\includegraphics[width=1.05\columnwidth]{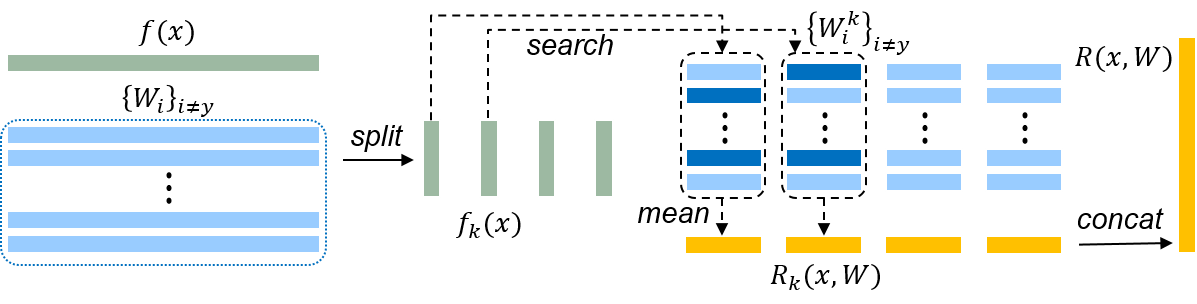}\vspace{-0.25cm}
	\caption{Illustration of high-level reconstruction.}\vspace{-0.4cm}
	\label{fig: high-level recon}
\end{figure}

To provide a better reconstruction of the extracted feature, we split the extracted feature $f(x)$ averagely into $S$ splits along the channel axis, denoted as $\{f_k(x)\}_{k=1}^S$ where $f_k(x) \in R^{d/S \times 1}$, i.e., concatenate all split features in $\{f_k(x)\}_{k=1}^S$ could obtain the original feature $f(x)$.
We also split each prototype into $S$ splits along the channel axis, where each split of prototypes is denoted as $\{W^k_i\}_{i \neq y}$, where $W^k_i \in R^{1 \times d/S}$. Then, the above nearest neighbor searching are conducted split-wisely between each $f_k(x)$ and $\{W^k_i\}_{i \neq y}$, outputting a split of reconstructed feature $R_k(x, W) \in R^{d/S \times 1}$, and finally the reconstructed feature $R(x, W) \in R^{d \times 1}$ is the concatenation of all splits of reconstructed features, as shown in Fig.~\ref{fig: high-level recon}. 
As the queried neighboring prototypes can be different across each split group $k$, the splitting operation can provide a closer reconstruction of $f(x)$ compared with directly applying the whole feature (Tab.~\ref{tab:recon_methods}).

%Specifically, to provide a finer reconstruction, we split the extracted feature $f(x)$ averagely into $S$ segments, denoted as $\{f_k(x)\}_{k=1}^S$, where $f_k(x) \in R^{d/S \times 1}$. This splitting operation is also carried along columns for whole prototype set $\widetilde{W}$, denoted as $\{\widetilde{W}^k\}_{k=1}^S$, where $\widetilde{W}^k \in R^{(N-1) \times d/S}$. 
%Then, within each segment group, we use extracted feature $f_k(x)$ to apply the nearest neighbor search over all $N-1$ base-class prototypes $\widetilde{W}^k$, and choose the top $m$ prototypes to form the neighboring prototype set, which is represented as

%%\vspace{-0.3cm}
%\begin{equation}
%\{\widetilde{W}^k_i | i \in T(s(\widetilde{W}^{k}, f_k(x)), m)\}_{k=1}^S
%\label{eq:proto_2}
%\end{equation}
%where $s(,)$ outputs $N-1$ cosine similarities, and $T(a, m)$ means get the top $m$ elements' indexes in vector $a$.
%Then, the reconstructed feature is calculated as the concatenation of all segment-wisely averaged neighboring prototypes as

%%\vspace{-0.3cm}
%\begin{equation}
%R(x, W) = concatenate(\{\frac{1}{m} \sum_{i=1}^{m} \widetilde{W}^k_i\}_{k=1}^S)
%\label{eq:proto_recon}
%\end{equation}
%where $concatenate()$ means concatenating vectors within the input set along the order of indexes. 

Then, we constrain the reconstructed feature to be close to $f(x)$ in the cosine similarity space with the loss

\vspace{-0.3cm}
\begin{equation}
L_{recon} = ||f^c(x) - R^c(x, W)||_2^2
\label{eq:L_recon}
\end{equation}
where $R^c(x, W)$ is the $L_2$ normalized $R(x, W)$.
%, and the squared Euclidean distance of $L_2$ normalized vectors are essentially equivalent to the cosine similarity. 
Note that by applying this loss, we are also trying to decompose $f(x)$ into the split prototypes, which can implicitly integrate the composition information of base classes into the feature, thus helping the in-domain FSL.

\vspace{-0.15cm}
\subsubsection{\textbf{Residual Calculation}}
\label{sec: residual calculation}

In experiments (Tab.~\ref{tab:recon_methods}), we find that for the best case that $f(x)$ could be improved for in-domain FSL, $L_{recon}$ remains about 0.11 to 0.25. Keeping enlarging the weight of $L_{recon}$ will largely decrease the performance, indicating the best case that the high-level reconstruction can reach, which verifies our assumption that every class has its character that could not be easily represented by high-level patterns from other classes.
By removing the high-level patterns from $f(x)$, we will get a discriminative residual feature which contains the discriminative information suitable for mid-level features to learn.

\begin{wrapfigure}{l}{2.2cm}
	%\vspace{-10pt}
	\includegraphics[width=3cm]{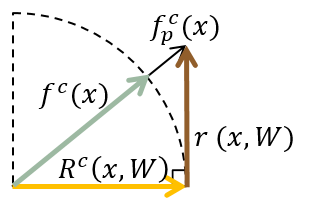}\\
	%\vspace{-20pt}
	\label{fig: recon_residual}
\end{wrapfigure}

%\begin{figure}
%\includegraphics[width=3cm]{recon_residual.png}\\
%%\vspace{-20pt}
%\label{fig: recon_residual}
%\end{figure}

%Therefore, we then calculate the residual term.
As both $f^c(x)$ and $R^c(x, W)$ are $L_2$ normalized, these vectors can be viewed to be distributed on a unit circle (left figure). Intuitively, the residual term and the high-level reconstructed term should not be representative of each other, which implies they should be orthogonal. Therefore, we prolong the $L_2$ normalized feature $f^c(x)$ to $f^c_p(x)$ by $f^c_p(x) = f^c(x) / cos \langle f^c(x), R^c(x, W) \rangle$, where the cosine value could be obtained as $1-L_{recon}/2$. Moreover, this prolonging could also stabilize the training. Thus, the residual is calculated as $r(x, W) = f^c_p(x) - R^c(x, W) \in R^{d \times 1}$.

%\vspace{-0.15cm}
\subsubsection{\textbf{Residual Prediction}}
\label{sec: mid-feature}

Then, to boost the discriminability of the mid-level features, we utilize the mid-level feature to predict the residual term. 
The predicted residual term is the weighted combination of multiple transformed mid-level features from a fixed mid-layer set $\{m_l(x)\}_{l=1}^L$ where $L$ is the number of total candidate mid-layers and $m_l(x) \in R^{d_l \times 1}$ is the mid-level feature for layer $l$.
For better understanding, we begin with the scenario where only one mid-layer $m_l(x)$ is used for the residual prediction, and the weighted combination of multiple layers will be included afterwards.

As a vector can be decomposed into its direction ($L_2$ normalized vector) and length ($L_2$ norm), we can re-write the residual term as

\vspace{-0.3cm}
\begin{equation}
r(x, W) = \frac{r(x, W)}{||r(x, W)||_2} \cdot ||r(x, W)||_2 = r^c(x, W) \cdot ||r(x, W)||_2
\label{eq:resi_direction_lenght}
\end{equation}

In practice, we find it is beneficial to predict the direction ($r^c(x, W)$) and the length $||r(x, W)||_2$ of the residual term separately.
For better understanding, we first introduce the prediction of the residual term's direction, which is similar to the prediction of its length.
Therefore, firstly our aim is to transform a mid-level feature $m_l(x)$ to predict $r^c(x, W)$.
As it is the mid-level feature instead of another high-level feature that is what we want, we should avoid learning another high-level feature by utilizing any deep and complex transformation network during the prediction. Therefore, we simply transform the mid-level feature by multiplying a matrix and adding a bias on it, which is calculated as
%Since the non-linearity of deep networks could transform a low-level feature to a higher-level feature layer by layer, we use a linear transformation layer to do the job, which is calculated as 

\vspace{-0.3cm}
\begin{equation}
\hat{r^c_l}(x, W) = \frac{W^r_l m_l(x) + b^r_l}{||W^r_l m_l(x) + b^r_l||_2}
\label{eq:mid2residual_direction}
\end{equation}
where $W^r_l \in R^{d \times d_l}$ and $b^r \in R^{d \times 1}$ are the weights and the biases for the transformation, and $d_l$ is the dimension of layer $l$. The prediction is also $L_2$ normalized to represent a direction, which simplifies the prediction.

Similarly, the prediction of the residual term's length $||r(x, W)||_2$ (a scalar) is also performed by a simple transformation as 

\vspace{-0.25cm}
\begin{equation}
\hat{r^s_l}(x, W) = W^s_l m_l(x) + b_l^s
\label{eq:mid2residual_length}
\end{equation}
where $W^s_l \in R^{d_l}$ and $b^s_l \in R$ are the corresponding parameters.

Given the predicted direction and length from $m_l(x)$, the predicted residual term can be represented as $\hat{r_l}_(x, W) = \hat{r^c_l}(x, W) \cdot \hat{r^s_l}(x, W) \in R^{d \times 1} $, and we design a loss as $||\hat{r_l^c}(x, W) - r^c(x, W)||_2^2 + \alpha  (\hat{r_l^s}(x, W) - ||r(x, W)||_2)^2$ to separately push the direction and length to be close to the residual term.

For the prediction from multiple mid-layers, we learn two weights for each layer's predicted direction and predicted length respectively, which are denoted as $a_l(x)$ and $a^s_l(x)$ and calculated as 

\vspace{-0.2cm}
\begin{equation}
\begin{aligned}
a_l(x) = \frac{t_l(m_l(x))}{\sum_{k=1}^L t_k(m_l(x))} ; \quad
a^s_l(x) = \frac{t^s_l(m_l(x))}{\sum_{k=1}^L t^s_k(m_l(x))}
\label{eq:weight}
\end{aligned}
\end{equation}

where $t_l()$ and $t^s_l(x)$ are implemented as two independent single-layer perceptrons appended to the corresponding mid-level feature $m_l(x)$ and each will output a scalar value.
With these layer-wise weights, we can simply use a fixed candidate mid-layer set to include all layers other than the first and the last layer.

Then, the weighted combination of the predicted direction and length from multiple mid-layers can be represented as

\vspace{-0.3cm}
\begin{equation}
\begin{aligned}
\hat{r^c}(x, W) &= \frac{\sum_{l=1}^L a_l(x) \cdot \hat{r^c_l}(x, W)}{||\sum_{l=1}^L a_l(x) \cdot \hat{r^c_l}(x, W)||_2} \\
\hat{r^s}(x, W) &= \sum_{l=1}^L a_l(x) \cdot \hat{r^s_l}(x, W)
\label{eq:combination}
\end{aligned}
\end{equation}

Finally, the prediction loss which separately pushes the direction and length to be close is calculated as 

\begin{figure}[t]
	\centering
	\hspace*{-0.3cm}
	\includegraphics[width=1.05\columnwidth]{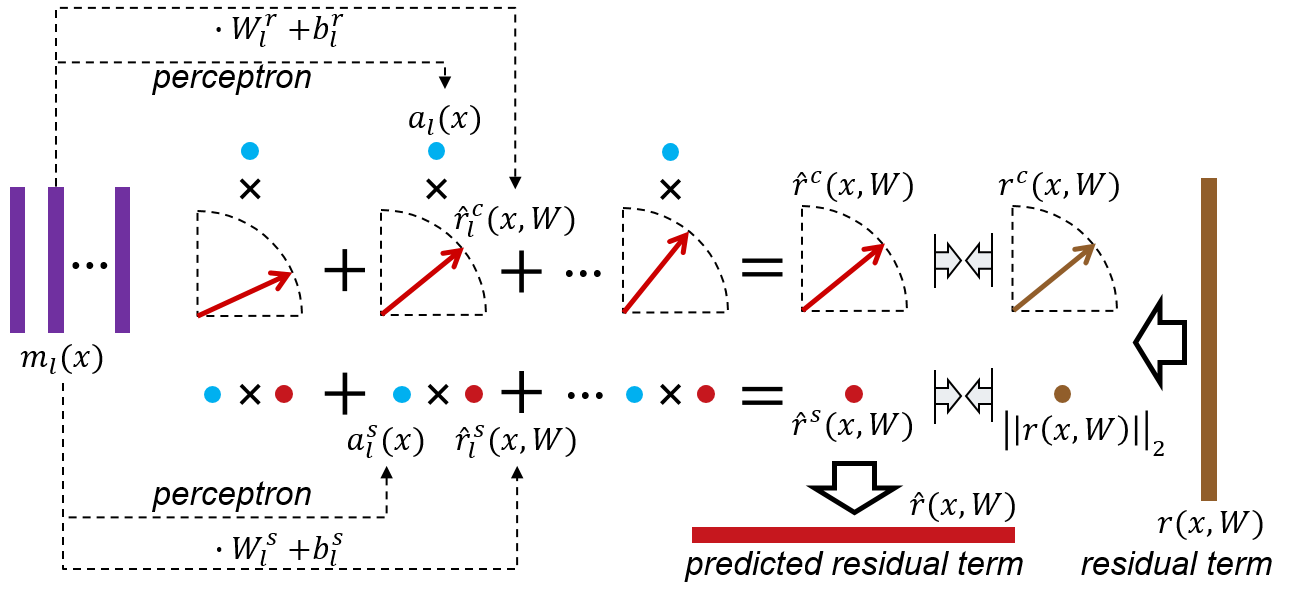}\vspace{-0.2cm}
	\caption{Illustration of residual prediction. 
		Each circle denotes a scalar, each arrow in a sector denotes a $L_2$ normalized vector, and blue circles denote the layer weights.
		Since a vector can be decomposed into its direction ($L_2$ normalized vector) and length ($L_2$ norm), we separately predict the residual term's direction $r^c(x, W)$ (brown arrow) and length $||r(x, W)||_2$ (brown circle). The predicted direction $\hat{r^c}(x, W)$ and length $\hat{r^s}(x, W)$ are the weighted combination of the each direction $\hat{r_l}(x, W)$ and $\hat{r^s_l}(x, W)$ which are transformed from each mid-layer $m_l(x)$, and are weighted by the layer specific weights $a_l(x)$ and $a^s_l(x)$ (blue circles).  }\vspace{-0.4cm}
	\label{fig: residual-prediction}
\end{figure}

\vspace{-0.3cm}
\begin{equation}
\small
L_{mid} = ||\hat{r^c}(x, W) - r^c(x, W)||_2^2 + \alpha  (\hat{r^s}(x, W) - ||r(x, W)||_2)^2
\label{eq:L_residual_1}
\end{equation}
where $r^c(x, W)$ and $||r(x, W)||_2$ is the direction and length of the residual term respectively, and $\alpha$ is a pre-defined hyper-parameter.
The first part of $L_{mid}$ is the direction prediction loss, and the second part is the length prediction loss. This process is shown in Fig.~\ref{fig: residual-prediction}.
%Note that while pushing the mid-level feature to predict the residual term, the linear layer is too simple to learn a discriminative feature on its own. Therefore, the discriminability originates from the mid-level features instead of the appended linear layer.
%Note that by minimizing $L_{mid}$, we are also encouraging the extracted feature to be decomposed into the high-level and mid-level feature, thus improving the extracted feature for in-domain FSL, which also validates the assumption that mid-level features are effective for describing the unique character of each class.

The final loss for base-class training is

\vspace{-0.2cm}
\begin{equation}
L = L_{cls} + \lambda_1 L_{recon} + \lambda_2 L_{mid}
\label{eq:final_loss}
\end{equation}
where $\lambda_1$ and $\lambda_2$ are predefined hyper-parameters.

%\vspace{-0.15cm}
\subsection{Novel-class Recognition}
\label{sec:novel-class}
Although we aim to boost mid-level features for distant-domain classes, this framework is also effective for in-domain FSL and CDFSL in near domains.
We provide two novel-class features under the same training framework, for both the distant- and in/near-domain settings respectively as shown in Fig.~\ref{fig: framework} (bottom), according to the quantitative domain distance measure by PAD~\cite{ganin2016domain}.

\vspace{-0.15cm}
\subsubsection{Distant-domain}
As all mid-level features within the candidate mid-feature set are improved but they are of different feature dimensions, we use the weighted concatenation of all mid-level features as the final feature as 

\vspace{-0.3cm}
\begin{equation}
F(x) = concatenate(\{a_l(x) \cdot m^c_l(x)\}_{l=1}^L)
\label{eq:F_cross}
\end{equation}
where $F(x)$ is the final feature in Eq.~\ref{eq:MN_P}, and $\{m^c_l(x)\}_{l=1}^L$ is the $L_2$ normalized candidate mid-feature set with the weight $a_l(x)$ in Eq.~\ref{eq:weight} for each layer. For clarity we denote this distant-domain feature as $F_a(x)$. The ablation study of each mid-feature is validated in Tab.~\ref{tab:ablation_cross_medical}.

\vspace{-0.2cm}
\subsubsection{In/near-domain}
The base-class training stated above is actually a pseudo-novel training strategy, which views the current sample $x$ as a pseudo-novel-class sample and views other classes as pseudo-base classes for it, providing simulated in-domain novel-class data.
As adequate information is provided for $L_{cls}$, the extracted pseudo-novel feature $f(x)$ can be viewed as the ground truth for such pseudo-novel training.
By encouraging the high-level reconstruction from pseudo-base classes, we are trying to predict the pseudo-novel feature $f(x)$  merely based on pseudo-base prototypes $\{W_i\}_{i \neq y}$.
By predicting the discriminative pseudo-novel residual feature, we are also encouraging the model to be able to predict the real discriminative residual feature for the (real) novel-class feature.
After such training, on base classes the model will be able to predict the discriminative pseudo-novel feature $f(x)$, merely based on the pseudo-base-class prototypes $\{W_i\}_{i \neq y}$ and mid-level features, which makes it able to predict the discriminative in/near-domain (real) novel-class feature, thus helping the in/near-domain FSL.

Therefore, we use all base-class prototypes to conduct the high-level reconstruction, and then the reconstructed feature together with the predicted residual term will be combined to be the final feature, which is calculated as 

\vspace{-0.2cm}
\begin{equation}
F(x) = R^c(x, W) + \hat{r^s}(x, W) \cdot \hat{r^c}(x, W)
\label{eq:F}
\end{equation}
For clarity we denote this in/near-domain feature as $F_b(x)$.

%Moreover, while minimizing both the $L_{recon}$ and the $L_{mid}$, we are actually trying to encoding the extracted feature with the high-level reconstructed feature $R^c(x, W)$ and the mid-level predicted residual $\hat{r^s}(x, W) \cdot \hat{r^c}(x, W)$. Therefore, the backbone-extracted feature is also improved and can be viewed as a simplified version

Finally, the nearest neighbor classification based on Eq.~\ref{eq:MN_P} will be conducted and obtain the final performance.

%%\vspace{-0.3cm}
\section{Experiments}

Extensive experiments are conducted under both distant-domain
%(on both few-shot image recognition task and few-shot video recognition task)
and in-domain FSL settings.
%(on both the pencil-paintings domain and the medical images). 
%We first introduce the datasets and implementation details, then compare our methods with state-of-the-art works, and finally show the ablation study of each module. 
Due to space limitation, more details are included in the supplementary material.

\vspace{-0.2cm}
\subsection{Datasets and Settings} 

\begin{table}[t]
	%\scriptsize
	\begin{center}
		\caption{Datasets and quantitative domain distances. }\vspace{-0.35cm}
		\label{tab:datasets}
		\resizebox{0.5\textwidth}{!}{
			\hskip-0.4cm
			\begin{tabular}{l|c|c|c|c}
				\hline\hline
				Dataset & Classes & Samples & Train/val/test class number & PAD
				\\
				\hline
				\textit{mini}ImageNet & 100 & 60,000 & 64 / 16 / 20 & 0.53 \\ 
				
				CUB-200-2011 & 200 & 11,788 & 100 / 50 / 50 & 0.60 \\
				
				Kinetics & 100 & 6,400 & 64 / 12 / 24 & - \\
				
				Pencil-paintings* & 36 & 21,600 & - / 16 / 20  & 1.80 \\
				
				TBN cell* & 3 & 363 & - / - / 3 & 1.80 \\
				
				Malaria cell* & 2 & 27,558 & - / - / 2 & 2.00 \\
				
				\hline\hline
		\end{tabular}}
	\end{center}\vspace{-0.5cm}
\end{table}

Datasets used for evaluation are summarized in Tab.~\ref{tab:datasets}, where \textit{val} denotes validation, and * denotes distant-domain datasets, including Pencil-paintings dataset~\cite{zhao2020domain} and two medical datasets\footnote{TBN Cell~\cite{pansombut2019convolutional}: http://mcs.sat.psu.ac.th/da- taset/dataset.zip  Malaria Cell~\cite{rajaraman2018pre}: https://lhn- cbc.nlm.nih.gov/publication/pub9932} (TBN Cell~\cite{pansombut2019convolutional} and Malaria Cell~\cite{rajaraman2018pre}).
\textit{mini}ImageNet~\cite{Vinyals2016Matching} is a subset of ImageNet~\cite{deng2009imagenet} and the pencil-paintings dataset contains novel classes of \textit{mini}ImageNet converted to the pencil-painting images. The CUB~\cite{wah2011caltech} dataset is a fine-grained dataset of birds, and the Kinetics dataset~\cite{zhu2018compound} contains video actions.
Some examples can be found in Fig.~\ref{fig: framework} (bottom).
For the distant-domain setting, the listed datasets are used as novel classes, and the base classes of \textit{mini}ImageNet are used as base classes~~\cite{DBLP:journals/corr/abs-1904-04232}. 
Following existing methods~~\cite{Vinyals2016Matching}, the mean accuracy (\%) and the 95\% confidence intervals of randomly generated $K$-way $n$-shot episodes from test sets (novel classes) will be reported.

\vspace{-0.25cm}
\subsection{Quantitative Measure of Domain Distances}
\label{sec: pad}

%Current works~\cite{DBLP:journals/corr/abs-1904-04232,tseng2020cross,zhao2020domain} choose cross-domain datasets merely by intuitively picking classes from different datasets.
%However, quantitative measure of domain distances is needed to distinguish cross-domain classes from in-domain classes.
We first utilize Proxy-A-Distance (PAD)~\cite{ganin2016domain,ben2007analysis} for quantitative domain distance measuring. 
As \textit{mini}ImageNet is used as the base classes for all the distant-domain evaluation, we calculate the PAD between all candidate domains' novel classes with base classes from \textit{mini}ImageNet.
Details are in supplementary materials, results are listed in the last column of  Tab.~\ref{tab:datasets}.

As \textit{mini}ImageNet's base and novel classes are from the same domain~\cite{Vinyals2016Matching}, the distance between these two set could be viewed as the lower bound for the PAD.
From Tab.~\ref{tab:datasets} we can see that the PAD of CUB is quite close to this lower bound, indicating this dataset should be used as the CDFSL dataset that is in the near domain of \textit{mini}ImageNet.
The pencil-paintings is the third distant domain. Unsurprisingly that it is closer than two medical datasets as it shares semantically similar classes with base classes of \textit{mini}ImageNet, but is much more distant than CUB. Therefore, we use this dataset as a distant domain. The two medical datasets are the furtherest datasets, with the Malaria cell dataset reaches the upper bound of PAD (2.0), so these two datasets are also selected as distant domains.

%\begin{table}[t]
%	%\scriptsize
%	\begin{center}
%		\caption{PAD based quantitative domain distances between \textit{mini}ImageNet base classes and others' novel classes. }%\vspace{-0.3cm}
%		\label{tab:proxy_distance}
%		\resizebox{0.45\textwidth}{!}{
%		\begin{tabular}{l|c|c|c|c|c}
%			\hline\hline
%			Evaluation & \textit{mini}ImageNet & CUB & Pencil & TBN cell & Malaria cell \\
%			\hline
%			%Accuracy (\%) & 63.33 & 65.00 & 89.17 & 95.00 & 100.00 \\
%			PAD & 0.53 & 0.60 & 1.57 & 1.80 & 2.00 \\
%			\hline\hline
%		\end{tabular}}
%	\end{center}%\vspace{-0.5cm}
%\end{table}

\vspace{-0.25cm}
\subsection{Implementation Details}

Due to the space limitation, detailed parameter settings are in the supplementary material.
%This method is implemented with TensorFlow~~\cite{abadi2016tensorflow} and is optimized with the Nesterov Momentum optimizer~~\cite{sutskever2013importance} with an initial learning rate of 0.01. 
For the CUB benchmark, we follow current works ~\cite{ye2018learning,triantafillou2017few,qiao2017few} to use the provided bounding boxes to crop the images. 
For the Kinetics dataset, to handle the temporal information, we uniformly sample 8 frames from the video,
%following the strategy of TSN~\cite{wang2016temporal}, 
and a temporal convolution layer with the kernel of 8 $\times$ 1 $\times$ 1 is appended to the backbone network. Following ~\cite{zhu2018compound}, we use the pre-trained weights from the ImageNet.
For all the above models, the mid-level feature maps of size $h_l \times w_l \times d_l$ ($t_l \times h_l \times w_l \times d_l$) is global averaged in all dimensions except the last one to obtain the mid-level feature.
%, which is then transposed to have the shape of $d_l \times 1$.
As all elements in extracted features are positive due to the ReLU~\cite{glorot2011deep} activation, to constrain prototypes in the same feature space as the extracted features, we also apply an $abs()$ function to FC parameters to use the absolute values.
For medical datasets, to preserve resolutions, we use raw images from ImageNet as base classes. 

\vspace{-0.2cm}
\subsection{Comparison with State-of-the-art}

\vspace{-0.05cm}
\subsubsection{Distant-domain Setting}

We report our performance under the distant-domain FSL setting in Tab.~\ref{tab:mini_dapn} and Tab.~\ref{tab:medical}, together with state-of-the-art works implemented by us.
%Note that as TBN and Malaria only have 3 and 2 classes respectively in all, we can only test the 3-way and 2-way classification performance on them respectively.
We denote the weighted concatenation of mid-features (i.e., $F_a()$) as $Ours_a$ in both tables.
All models are trained on base classes of \textit{mini}ImageNet. We use ResNet12~\cite{he2016deep} as the backbone network and use features of the third block and second block to form the mid-layer candidate set. From these tables we can see that our method outperforms the state-of-the-art methods on all three datasets.

\begin{table}[t]
	%\tiny
	\begin{center}
		\caption{Distant-domain performance on pencil-paintings. }\vspace{-0.35cm}
		\label{tab:mini_dapn}
		\resizebox{0.32\textwidth}{!}{
			\begin{tabular}{l|c|c}
				\hline\hline
				Method & 5-way 1-shot & 5-way 5-shot
				\\
				\hline
				MatchingNet~\cite{Vinyals2016Matching} & $23.35 \pm 0.64$ & $32.42 \pm 0.55$ \\
				RelationNet~\cite{yang2018learning} & $23.87 \pm 0.82$ & $33.29 \pm 0.96$ \\
				PPA~\cite{qiao2017few} & $23.86 \pm 0.42$ & $33.74 \pm 0.41$ \\
				SGM~\cite{wang2018low} & $23.49 \pm 0.29$ & $32.67 \pm 0.32$ \\
				ProtoNet~\cite{snell2017prototypical} & $23.23 \pm 0.32$ & $32.92 \pm 0.41$ \\
				MetaOptNet~\cite{lee2019meta} & $24.53 \pm 0.20$ & $33.23 \pm 0.63$ \\
				Baseline++~\cite{DBLP:journals/corr/abs-1904-04232} & $24.06 \pm 0.46$ & $32.74 \pm 0.81$ \\
				LFT+GNN~\cite{tseng2020cross} & $ 27.02 \pm 0.43 $  & $ 34.28 \pm 0.43 $ \\
				DAPN~\cite{zhao2020domain} & $27.25 \pm 0.25$ & $37.45 \pm 0.25$ \\
				\hline
				Cosine Classifier & $ 28.19 \pm 0.24 $ & $37.21 \pm 0.37$ \\
				Ours$_a$ & $\textbf{29.45} \pm \textbf{0.22}$ & $ \textbf{40.38} \pm \textbf{0.35}$ \\
				\hline\hline
		\end{tabular}}
	\end{center}\vspace{-0.5cm}
\end{table}

\begin{table}[t]
	%\scriptsize
	%\footnotesize
	\begin{center}
		\caption{Distant-domain performance on medical datasets. }\vspace{-0.35cm}
		\label{tab:medical}
		\resizebox{0.50\textwidth}{!}{
			\hskip-0.65cm
			\begin{tabular}{l|c|c|c|c}
				\hline\hline
				\multirow{2}{*}{\tabincell{c}{Method}} & \multicolumn{2}{c|}{TBN cell (3-way)} & \multicolumn{2}{c}{Malaria cell (2-way)} \\
				\cline{2-5}
				& 1-shot & 5-shot & 1-shot & 5-shot
				\\
				\hline
				Pixel & $44.03 \pm 0.27$ & $51.95 \pm 0.38$ & $53.01 \pm 0.40$ & $53.79 \pm 0.53$ \\
				Random Init & $48.38 \pm 0.33$ & $55.87 \pm 0.44$ & $52.75 \pm 0.36$ & $55.78 \pm 0.58$ \\
				MatchingNet~\cite{Vinyals2016Matching} & $44.40 \pm 0.31$ & $60.52 \pm 0.43$ & $53.92 \pm 0.38$ & $57.65 \pm 0.54$  \\
				DAPN~\cite{zhao2020domain} & $ 54.18 \pm 0.38 $ & $ 64.56 \pm 0.29 $ & $ 55.22 \pm 0.40 $ & $63.88 \pm 0.37$ \\
				ProtoNet~\cite{snell2017prototypical} & $59.56 \pm 0.28$ & $66.48 \pm 0.39$ & $58.12 \pm 0.37$ & $67.68 \pm 0.55$ \\
				Baseline++~\cite{DBLP:journals/corr/abs-1904-04232} & $56.89 \pm 0.32$ & $66.25 \pm 0.40$ & $60.47 \pm 0.37$ & $71.35 \pm 0.48$\\
				MatchingNet+~\cite{chen2020new} & $51.54 \pm 0.31$ & $62.57 \pm 0.39$ & $59.97 \pm 0.39$ & $64.47 \pm 0.52$  \\
				ProtoNet+~\cite{chen2020new} & $60.07 \pm 0.28$ & $66.56 \pm 0.38$ & $59.85 \pm 0.38$ & $70.06 \pm 0.52$  \\
				LFT+GNN~\cite{tseng2020cross} & $ 54.20 \pm 0.39 $  & $ 67.13 \pm 0.31 $ & $ 62.54 \pm 0.52 $ & $ 74.51 \pm 0.38 $ \\
				Multi-level~\cite{huang2019all} & $61.71 \pm 0.28$ & $68.95 \pm 0.40$ & $60.86 \pm 0.39$ & $72.60 \pm 0.54$ \\
				\hline
				Cosine Classifier & $60.65 \pm 0.28$ & $68.96 \pm 0.36$ & $59.16 \pm 0.35$ & $70.41 \pm 0.39$ \\
				Ours$_a$ & $\textbf{64.12} \pm \textbf{0.26}$ & $\textbf{72.88} \pm \textbf{0.36}$ & $\textbf{63.82} \pm \textbf{0.41}$ & $\textbf{76.94} \pm \textbf{0.32}$ \\
				
				\hline\hline
		\end{tabular}}
	\end{center}\vspace{-0.4cm}
\end{table}

\begin{table}[t]
	%\tiny
	\begin{center}
		\caption{Cross-domain performance on near domains (CUB). }\vspace{-0.35cm}
		\label{tab:mini_cub}
		\resizebox{0.32\textwidth}{!}{
			\begin{tabular}{l|c|c}
				\hline\hline
				Method & 5-way 1-shot & 5-way 5-shot
				\\
				\hline
				MatchingNet~\cite{Vinyals2016Matching} & $35.89 \pm 0.51$ & $51.37 \pm 0.77$ \\
				RelationNet~\cite{yang2018learning} & $42.44 \pm 0.77$ & $57.77 \pm 0.69$ \\
				ProtoNet~\cite{snell2017prototypical} & $-$ & $62.02 \pm 0.70$ \\
				Baseline++~\cite{DBLP:journals/corr/abs-1904-04232} & $-$ & $65.57 \pm 0.70$ \\
				DAPN~\cite{zhao2020domain} & $47.47 \pm 0.75$ & $66.98 \pm 0.68$ \\
				Neg-Cosine~\cite{liu2020negative} & $ - $ & $67.03 \pm 0.76$ \\
				\hline
				Cosine Classifier & $ 45.97 \pm 0.29 $ & $ 65.92 \pm 0.24 $ \\
				Ours$_b$ & $\textbf{48.81} \pm \textbf{0.30}$ & $ \textbf{67.04} \pm \textbf{0.61}$ \\
				\hline\hline
		\end{tabular}}
	\end{center}\vspace{-0.5cm}
\end{table}

\vspace{-0.1cm}
\subsubsection{In/near-domain Setting}

To compare our method with current works in the in/near-domain setting, we report the performance of ours and that of others in Tab.~\ref{tab:CUB}, \ref{tab:miniImagenet} and \ref{tab:Kinetics} under the in-domain setting, and report the comparison of CDFSL in near domains in Tab.~\ref{tab:mini_cub}.
The feature provided for the in/near-domain setting (i.e., $F_b()$) is denoted as $Ours_b$.
For the \textit{mini}ImageNet, we include experiments of the \textit{trainval} settings~\cite{qiao2017few,rusu2019meta,lee2019meta}, where the validation set is included during training.
In Tab.\ref{tab:Kinetics}, results of current works are directly obtained from \cite{zhu2018compound} (without std).
%The backbone for CUB is ResNet10, for Kinetics is ResNet12.
As shown in these tables, we outperform current works on CUB and Kinetics, and achieve comparable performance on \textit{mini}ImageNet.

\begin{table}[t]
	%\scriptsize
	\begin{center}
		\caption{In-domain image FSL on CUB (ResNet10).}\vspace{-0.35cm}
		\label{tab:CUB}
		\resizebox{0.34\textwidth}{!}{
			\begin{tabular}{l|c|c}
				\hline\hline
				Method & 5-way 1-shot & 5-way 5-shot
				\\
				\hline
				MatchingNet~\cite{Vinyals2016Matching} & $61.16 \pm 0.89$ & $72.86 \pm 0.70$ \\
				ProtoNet~\cite{snell2017prototypical} & $51.31 \pm 0.91$ & $70.77 \pm 0.69$ \\
				MAML~\cite{finn2017model} & $55.92 \pm 0.95$ & $72.09 \pm 0.76$ \\
				RelationNet~\cite{yang2018learning} & $62.45 \pm 0.98$ & $76.11 \pm 0.69$ \\
				DEML+MetaSGD~\cite{zhou2018deep} & $66.95 \pm 1.06$ & $77.1 \pm 0.78$ \\
				ResNet18+TriNet~\cite{chen2018semantic} & $69.61 \pm 0.46$ & $84.10 \pm 0.35$ \\
				MAML++~\cite{antoniou2019learning} & $67.48 \pm 1.44$ & $83.80 \pm 0.35$ \\
				SCA+MAML++~\cite{antoniou2019learning} & $70.33 \pm 0.78$ & $85.47 \pm 0.40$ \\
				S2M2~\cite{mangla2019charting} & $72.40 \pm 0.34$ & $86.22 \pm 0.53$ \\
				CFA~\cite{hu2019weakly} & $73.90 \pm 0.80$ & $86.80 \pm 0.50$ \\
				AssoAlign~\cite{afrasiyabi2020associative} & $74.22 \pm 1.09$ & $88.65 \pm 0.55$ \\
				DeepEmb~\cite{zhang2020deepemd} & $75.65 \pm 0.83$ & $88.69 \pm 0.50$ \\
				\hline
				Cosine Classifier & $ 71.37 \pm 0.25 $ & $ 86.57 \pm 0.46 $ \\
				Ours$_b$ & $\textbf{77.65} \pm \textbf{0.26}$ & $ \textbf{88.83} \pm \textbf{0.48}$ \\
				\hline\hline
		\end{tabular}}
	\end{center}\vspace{-0.5cm}
\end{table}

\begin{table}[t]
	%\scriptsize
	\begin{center}
		\caption{In-domain image FSL on \textit{mini}ImageNet.}\vspace{-0.35cm}
		\label{tab:miniImagenet}
		\resizebox{0.38\textwidth}{!}{
			\begin{tabular}{l|c|c}
				\hline\hline
				Method & 5-way 1-shot & 5-way 5-shot
				\\
				\hline
				%Meta-LSTM~\cite{ravi2016optimization} & $43.44 \pm 0.77$ & $60.60 \pm 0.71$ \\
				MatchingNet~\cite{Vinyals2016Matching} & $46.56 \pm 0.84$ & $55.31 \pm 0.73$ \\
				ProtoNet~\cite{snell2017prototypical} & $49.42 \pm 0.78$ & $68.20 \pm 0.66$ \\
				MAML~\cite{finn2017model} & $48.70 \pm 1.84$ & $63.11 \pm 0.92$ \\
				RelationNet~\cite{yang2018learning} & $50.44 \pm 0.82$ & $65.32 \pm 0.70$ \\
				%AgileNet~\cite{ghasemzadeh2018agilenet} & $58.23 \pm 0.10$ & $71.39 \pm 0.10$ \\
				%DEML+MetaSGD~\cite{zhou2018deep} & $58.49 \pm 0.91$ & $71.28 \pm 0.69$ \\
				%Dynamic FS~\cite{gidaris2018dynamic} & $56.20 \pm 0.86$ & $73.00 \pm 0.64$ \\
				Dynamic FS~\cite{gidaris2018dynamic} & $55.45 \pm 0.89$ & $70.13 \pm 0.68$ \\
				SNAIL~\cite{mishra2017simple} & $55.71 \pm 0.99$ & $ 68.88 \pm 0.92 $ \\
				TADAM~\cite{oreshkin2018tadam} & $58.50 \pm 0.30$ & $76.70 \pm 0.30$ \\
				PPA (\textit{trainval})~\cite{qiao2017few} & $59.60 \pm 0.41$ & $73.74 \pm 0.19$ \\
				LEO (\textit{trainval})~\cite{rusu2019meta} & $ 61.76 \pm 0.08 $ & $ 77.59 \pm 0.12 $ \\
				DCO~\cite{lee2019meta} & $62.62 \pm 0.61 $ & $ 78.63 \pm 0.46$ \\
				DCO (\textit{trainval})~\cite{lee2019meta} & $64.09 \pm 0.62 $ & $ 80.00 \pm 0.45$ \\
				%Neg-Cosine~\cite{liu2020negative} & 
				MetaOptNet~\cite{lee2019meta} & $ 62.64 \pm 0.61 $ & $78.63 \pm 0.46$ \\
				MetaOptNet (\textit{trainval})~\cite{lee2019meta} & $ 64.09 \pm 0.62 $ & $80.00 \pm 0.45$ \\
				%TransMatch (semi-supervised)~\cite{yu2020transmatch} & $ 63.02 \pm 1.07$ & $81.19 \pm 0.59$ \\
				\hline
				Cosine Classifier (ResNet10) & $55.97 \pm 0.26$ & $ 74.95 \pm 0.24 $ \\
				Ours$_b$ (ResNet10) & $ 62.01 \pm 0.27 $ & $ 77.49 \pm 0.63  $ \\
				Ours$_b$ (\textit{trainval}, ResNet10) & $ 63.38 \pm 0.69 $ & $ 78.72 \pm 0.65 $ \\
				\hline
				Cosine Classifier (ResNet12) & $56.26 \pm 0.28$ & $ 74.97 \pm 0.24 $ \\
				Ours$_b$ (ResNet12) & $ 63.06 \pm 0.76 $ & $ 77.82 \pm  0.58 $ \\
				Ours$_b$ (\textit{trainval}, ResNet12) & $ \textbf{64.19} \pm \textbf{0.78} $ & $ \textbf{80.01} \pm \textbf{0.76} $ \\
				\hline\hline
		\end{tabular}}
	\end{center}\vspace{0.05cm}
	
	\begin{center}
		\caption{In-domain video FSL on Kinetics (ResNet12).}\vspace{-0.35cm}
		\label{tab:Kinetics}
		\resizebox{0.35\textwidth}{!}{
			\begin{tabular}{l|c|c}
				\hline\hline
				Method & \tabincell{c}{5-way 1-shot} & \tabincell{c}{5-way 5-shot}
				\\
				\hline
				RGB w/o mem & $28.7$ & $48.6$ \\
				Flow w/o mem & $24.4$ & $33.1$ \\
				LSTM(RGB) w/o mem & $28.9$ & $49.0$ \\
				Nearest-finetune & $48.2$ & $62.6$ \\
				Nearest-pretrain & $51.1$ & $68.9$ \\
				Matching Network~\cite{Vinyals2016Matching} & $53.3$ & $74.6$ \\
				MAML~\cite{finn2017model} & $54.2$ & $75.3$ \\
				Plain CMN~\cite{kaiser2017learning} & $57.3$ & $76.0$ \\
				LSTM-cmb & $57.6$ & $76.2$ \\
				CMN~\cite{zhu2018compound} & $60.5$ & $78.9$ \\
				
				\hline
				Cosine classifier & $ 62.13 \pm 0.27 $  & $ 77.81 \pm 0.63 $ \\
				Ours$_b$ & $\textbf{64.63} \pm \textbf{0.64}$ & $ \textbf{79.10} \pm \textbf{0.77} $ \\
				\hline\hline
		\end{tabular}}
	\end{center}\vspace{-0.5cm}
\end{table}

\vspace{-0.15cm}
\subsection{Ablation Study}

For better understanding, we report the ablation studies first for the in-domain FSL and then for the distant-domain FSL.

\vspace{-0.15cm}
\subsubsection{In-domain Setting}
\label{sec: component}

\begin{figure*}
	\begin{minipage}[t]{0.48\textwidth}
		\centering
		\makeatletter\def\@captype{table}\makeatother\caption{Ablation study by distant-domain 1-shot testing.}
		\begin{tabular}{cccc} 
			\label{tab:ablation_cross_medical}
			\resizebox{1.0\textwidth}{!}{
				\hskip-1.5cm
				\begin{tabular}{c|c|c|c|c|c}
					\hline\hline
					Case & Method & \makecell[c]{Forth Block \\ (Final-layer)} & \makecell[c]{Third Block \\ (Mid-layer)} & \makecell[c]{Second Block \\ (Mid-layer)} & \makecell[c]{Concatenation of \\ mid-features (Ours$_a$)} \\
					\hline
					\multirow{4}{*}{\tabincell{c}{Pencil\\ResNet12\\5-way 1-shot}} & Cosine & $28.19 \pm 0.24$ & $28.35 \pm 0.23$ & $27.21 \pm 0.21$ & $27.95 \pm 0.23$ \\
					& +abs & $28.22 \pm 0.21$ & $28.84 \pm 0.21$ & $27.95 \pm 0.22$ & $28.89 \pm 0.22$ \\
					& + $L_{recon}$ & $28.29 \pm 0.22$ & $28.12 \pm 0.24$ & $27.61 \pm 0.24$ & $28.12 \pm 0.23$ \\
					& + $L_{mid}$ & $28.09 \pm 0.21$ & $29.21 \pm 0.25$ & $28.13 \pm 0.22$ & $\textbf{29.45} \pm \textbf{0.22}$ \\
					\hline
					\multirow{4}{*}{\tabincell{c}{TBN Cell\\ResNet12\\3-way 1-shot}} & Cosine & $60.25 \pm 0.28$ & $60.27 \pm 0.29$ & $60.74 \pm 0.27$ & $60.55 \pm 0.28$ \\
					& +abs & $62.22 \pm 0.32$ & $62.82 \pm 0.29$ & $62.05 \pm 0.31$ & $61.78 \pm 0.28$ \\
					& + $L_{recon}$ & $58.66 \pm 0.31$ & $61.09 \pm 0.30$ & $60.57 \pm 0.29$ & $60.95 \pm 0.27$ \\
					& + $L_{mid}$ & $59.61 \pm 0.29$ & $62.76 \pm 0.27$ & $63.98 \pm 0.27$ & $\textbf{64.12} \pm \textbf{0.26}$ \\
					\hline
					\multirow{4}{*}{\tabincell{c}{Malaria Cell\\ResNet12\\2-way 1-shot}} & Cosine & $59.16 \pm 0.35$ & $61.04 \pm 0.37$ & $59.66 \pm 0.38$ & $59.90 \pm 0.37$ \\
					& +abs & $59.50 \pm 0.32$ & $60.40 \pm 0.36$ & $58.86 \pm 0.39$ & $59.68 \pm 0.38$ \\
					& + $L_{recon}$ & $58.83 \pm 0.34$ & $61.40 \pm 0.39$ & $59.97 \pm 0.39$ & $60.79 \pm 0.38$ \\
					& + $L_{mid}$ & $61.63 \pm 0.36$ & $63.29 \pm 0.35$ & $63.14 \pm 0.41$ & $\textbf{63.82} \pm \textbf{0.41}$ 
					\\
					\hline\hline
			\end{tabular}}
		\end{tabular}
	\end{minipage} \hspace{1.0cm}
	\begin{minipage}[t]{0.4\textwidth}
		\centering
		\makeatletter\def\@captype{table}\makeatother\caption{Ablation study under the in-domain setting.}
		\label{tab:ablation}
		\resizebox{1.0\textwidth}{!}{
			%\hskip1.5cm
			\begin{tabular}{c|c|c|c}
				\hline\hline
				Study Case & Method  & 5-way 1-shot (\%) & 5-way 5-shot (\%)
				\\
				\hline
				\multirow{4}{*}{\tabincell{c}{CUB \\ ResNet10}} & cosine classifier & $ 71.37 \pm 0.25 $ & $ 86.57 \pm 0.46 $ \\
				& + abs & $ 72.84 \pm 0.26 $ & $ 86.99 \pm 0.23 $\\
				& + $L_{recon}$ & $ 76.47 \pm 0.27 $ & $ 87.22 \pm 0.24 $\\
				& + $L_{mid}$ & $\textbf{77.65} \pm \textbf{0.26}$ & $ \textbf{88.23} \pm \textbf{0.48} $ \\
				%\cline{2-4}
				%& backbone & $ 77.58 \pm 0.65 $ & $ 88.11 \pm 0.25 $ \\
				\hline
				\multirow{4}{*}{\tabincell{c}{\textit{mini}ImageNet \\ ResNet10 }} & cosine classifier & $ 55.97 \pm 0.26 $ & $ 74.95 \pm 0.24 $ \\
				& + abs & $ 57.83 \pm 0.26 $ & $ 75.25 \pm 0.71 $\\
				& + $L_{recon}$ & $ 59.15 \pm 0.25 $ & $ 75.73 \pm 0.83 $\\
				& + $L_{mid}$ & $\textbf{62.01} \pm \textbf{0.27}$ & $ \textbf{77.49} \pm \textbf{0.63} $ \\
				%\cline{2-4}
				%& backbone & $61.20 \pm 0.62$ & $ 77.44 \pm 0.64 $ \\
				\hline
				\multirow{4}{*}{\tabincell{c}{Kinetics \\ ResNet12 }} & cosine classifier & $ 62.13 \pm 0.27 $ & $ 77.81 \pm 0.63 $ \\
				& + abs & $ 62.67 \pm 0.26 $ & $ 78.27 \pm 0.61 $\\
				& + $L_{recon}$ & $ 63.59 \pm 0.25 $ & $ 78.91 \pm 0.60 $\\
				& + $L_{mid}$ & $\textbf{64.63} \pm \textbf{0.64}$ & $ \textbf{79.10} \pm \textbf{0.77} $ \\
				%\cline{2-4}
				%& backbone & $64.01 \pm 0.52$ & $ 78.90 \pm 0.77 $ \\
				\hline\hline
		\end{tabular}}
	\end{minipage}
\end{figure*}

The performance of models implemented with different modules is in Tab.~\ref{tab:ablation}. Each row in the second column represents the model with modules of all the above rows plus the module in the current row. 
%The improvement of each module is shown in the third and forth columns of the corresponding row. 
For the cosine classifier, $+abs$, and $+L_{recon}$, we use the backbone extracted feature as the final feature (i.e., $F=f$ in Eq.~\ref{eq:MN_P}). For $+L_{mid}$, we use $F_b()$ as the final feature.
%, which corresponds to the final performance in Tab.~\ref{tab:CUB}, \ref{tab:miniImagenet} and \ref{tab:Kinetics}.
%We also report the performance evaluated by the backbone extracted feature of the final model ($+L_{mid}$) in the last row of each case.

From this table we can see that each module contributes to the performance respectively. $abs$ contributes because all the prototypes are constrained to be positive, which is the same as the extracted features, simplifying the training. 
$L_{recon}$ also improves the performance because by encouraging the pseudo-novel feature close to the reconstructed feature, the network is pushed to learn the composition of base-class prototypes.
%, making the extracted feature implicitly encoded with base-class prototypes.
Moreover, we can see that $L_{recon}$ contributes the most to CUB by around 4\%, exceeding that to all other datasets. On the other hand, for $L_{mid}$, on CUB this term contribute less than that on \textit{mini}ImageNet. This is because CUB is a fine-grained dataset for birds. As nearly all birds contain similar high-level patterns (e.g., wings, beak), the high-level feature reconstruction on CUB will be much easier than that on \textit{mini}ImageNet. In \textit{mini}ImageNet, various classes exist, such as dogs, cars, ships, which means these classes may not contains as many over-lapped high-level patterns as that on CUB, making the high-level feature reconstruction harder. Therefore, $L_{recon}$ promotes more on CUB. Moreover, it can also explain why $L_{mid}$ promotes more on \textit{mini}ImageNet than CUB: As there are larger residual terms on \textit{mini}ImageNet that could not be well reconstructed by prototypes, the mid-level prediction helps more. 

\vspace{-0.15cm}
\subsubsection{Distant-domain Setting}

The ablation study is in Tab.~\ref{tab:ablation_cross_medical}, with the feature of the forth (i.e., final), the third, the second block and the concatenation of the third and second block (i.e., $F_a()$). We can see that almost all forth blocks' features are outperformed by that of the third blocks, which is consistent with the study~\cite{yosinski2014transferable} that mid-level features can be more transferable than final-layer's feature, and verifies the choice of distant domain datasets. For this challenging setting, methods that helps the in-domain FSL cannot show significant improvements now, because by fitting the base classes (especially $+L_{recon}$), the model is likely to learn more about the domain-specific information which may harm the transferability.
Meanwhile, by applying $L_{mid}$, we can see a clear improvement of mid-level features over baselines and the forth layer features, which verifies the our motivation: boosting the discriminability of mid-level features by the residual-prediction task.

\vspace{-0.15cm}
\subsubsection{Comparison of high-level reconstruction}

%To verify the choice of high-level feature reconstruction and that the residual term exists, 
We trained our model with $L_{recon}$ on \textit{mini}ImageNet with different splits in Tab.~\ref{tab:recon_methods}, with the optimal performance and the $L_{recon}$ value. 
We can see by splitting the extracted feature and prototypes, the reconstructed feature gets closer to the extracted feature ($L_{recon}$ decreases as splits increase), leading to the best feature implicitly encoded with base-class prototypes (setting splits to 4, obtaining the accuracy at 59.15).
Also, we can see that for the best improved feature, $L_{recon}$ remains larger than 0.1, which coarsely corresponds to an angle of 20$^{\circ}$ in the degree measure, verifying the existence of residual terms.

\begin{table}
	\begin{center}
		\caption{Hyper-parameters of high-level feature reconstruction. Experiments are conducted on \textit{mini}ImageNet.} \vspace{-0.35cm}
		\label{tab:recon_methods}
		\resizebox{0.4\textwidth}{!}{
			\begin{tabular}{c|c|c|c}
				\hline\hline
				Method  & Segments & 5-way 1-shot (\%) & $L_{recon}$
				\\
				\hline
				without $L_{recon}$ & - & $57.83 \pm 0.26$ & - 
				\\
				\hline
				\multirow{4}{*}{\tabincell{c}{Average over queried}} & 1 & $58.25 \pm 0.28$ & $0.19$ \\
				%& 2 & $58.07 \pm 0.26$ & $0.17$ \\
				& 4 & $\textbf{59.15} \pm \textbf{0.25}$ & $0.16$ \\
				%& 8 & $58.78 \pm 0.27$ & $0.15$ \\
				& 16 & $58.94 \pm 0.26$ & $0.14$ \\
				%& 32 & $59.02 \pm 0.25$ & $0.12$ \\
				& 64 & $58.21 \pm 0.26$ & $0.11$ \\
				\hline\hline
		\end{tabular}}
	\end{center}\vspace{-0.5cm}
\end{table}

\vspace{-0.15cm}
\subsection{Visualization}

\begin{figure}[t]
	\centering\includegraphics[width=1.0\columnwidth, height=0.35\columnwidth]{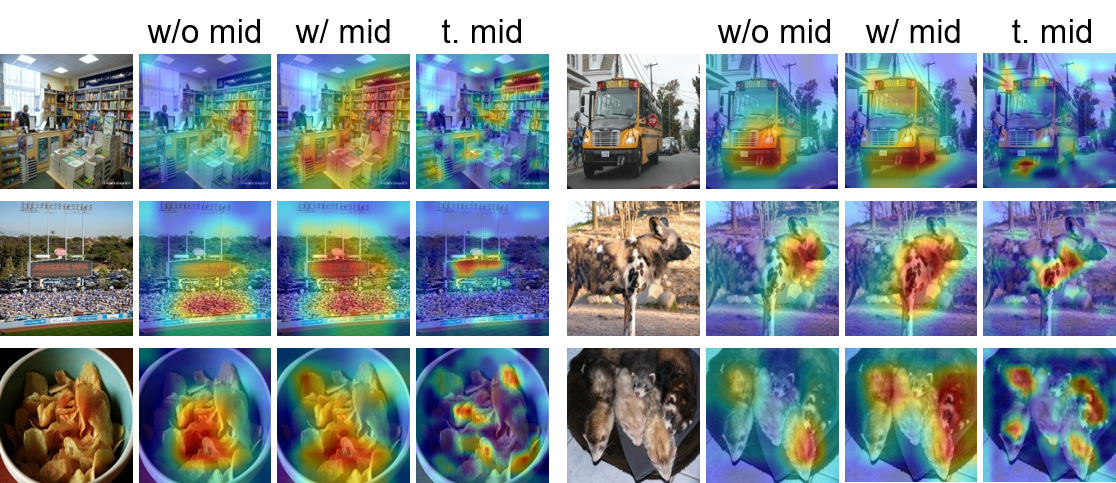}\vspace{-0.2cm}
	\caption{Visualization of the final-layer feature maps when trained with (w/) and without (w/o) $L_{mid}$, and the transformed mid-layer feature maps (t. mid). w/ mid cover more activated regions than w/o mid, where the difference can coarsely correspond to that of t. mid.}\vspace{-0.6cm}
	\label{fig: visualization}
\end{figure}

To verify the contribution of mid-level patterns, we visualize the activated regions on novel classes of \textit{mini}ImageNet in Fig.~\ref{fig: visualization}. As the high-level and mid-level representations are already implicitly encoded in the extracted feature, we visualize the final-layer feature maps of models trained with (w/) and without (w/o) the $L_{mid}$, together with the  transformed mid-layer feature maps (t. mid), by means of summing up and resizing all the feature maps~\cite{zhou2016learning}. 
We can see the model trained without $L_{mid}$ covers less activated regions than that trained with $L_{mid}$, indicating regions that are unable to be described by the baseline method are now better described by our model. The difference in the covered activated regions coarsely corresponds to the activated regions of the transformed mid-layer feature maps, which verifies the contribution of mid-level patterns.

\vspace{-0.15cm}
\section{Conclusion}

To learn transferable and discriminative mid-level features for the distant-domain FSL, we proposed a residual-prediction task consisting of the high-level feature reconstruction and the mid-level residual prediction, which consistently achieves state-of-the-art performance and better. Extensive experiments on both the distant- and in-domain settings including image recognition and video recognition show the rationale and the insights of the proposed method.

\vspace{-0.15cm}
\section{Acknowledgments}
This work is partially supported by Key-Area Research and Development Program of Guangdong Province under contact No.2019B0101-53002, and grants from the National Natural Science Foundation of China under contract No. 61825101 and No. 62088102.

\bibliographystyle{ACM-Reference-Format}
%\balance
\bibliography{zoilsen}

\end{document}